\documentclass{article}

\usepackage{arxiv}

\usepackage[utf8]{inputenc} 
\usepackage[T1]{fontenc}    
\usepackage{hyperref}       
\usepackage{url}            
\usepackage{booktabs}       
\usepackage{amsfonts}       
\usepackage{nicefrac}       
\usepackage{microtype}      
\usepackage{lipsum}
\usepackage{svg}
\usepackage{graphicx}
\graphicspath{ {./images/} }
\usepackage{amsmath} 
\usepackage{pdflscape}
\usepackage{xcolor}
\usepackage[normalem]{ulem}
\usepackage{comment}
\usepackage{longtable}
\usepackage{tabularx}
\usepackage{caption} 
\usepackage{subcaption}
\usepackage{longtable}
\usepackage{booktabs}
\usepackage[utf8]{inputenc}
\usepackage{changepage}

\usepackage{array}
\usepackage{float}
\newlength{\extralength}
\newlength{\fulllength}
\newcolumntype{C}{>{\centering\arraybackslash}X}

\title{YOLO Network For Defect Detection In Optical lenses}

\author{Habib Yaseen\\[1ex]
\begin{minipage}[t]{0.90\textwidth}
\centering
\scriptsize School of Computing and Engineering, University of Huddersfield, Queensgate, Huddersfield HD1 3DH, UK; \\
\textsuperscript{*}Correspondence: U2391139@unimail.hud.ac.uk;
\end{minipage}}

\begin{document}

\maketitle
\begin{abstract}
Mass-produced optical lenses often exhibit defects that alter their scattering properties and compromise quality standards. Manual inspection is usually adopted to detect defects, but it is not recommended due to low accuracy, high error rate and limited scalability. To address these challenges, this study presents an automated defect detection system based on the YOLOv8 deep learning model. A custom dataset of optical lenses, annotated with defect and lens regions, was created to train the model. Experimental results obtained in this study reveal that the system can be used to efficiently and accurately detect defects in optical lenses. The proposed system can be utilized in real-time industrial environments to enhance quality control processes by enabling reliable and scalable defect detection in optical lens manufacturing.
\end{abstract}


\section{Introduction}
An optical lens is a transparent optical component designed to focus or disperse light by refraction. Lenses are integral in systems such as cameras, microscopes, and human vision, where they project sharp images onto surfaces or sensors by bending light rays to a focal point \cite{Ochmas2023Examining}. Proper design and quality of lenses are critical for their functionality, as they directly affect image clarity and optical performance of optical systems used in industries ranging from photography and microscopy to astronomy and telecommunications \cite{tang2023improved}.
However, owing to the uncertainty associated with environmental variables and process parameters, the mass production of lenses inevitably results in the occurrence of several defects that tend to affect their optical properties \cite{maschmeyer1983precision}. Therefore, there exists a need for development of a sophisticated defect-detection method.
Until now, much of the defect detection work of lenses has been performed manually, that leads to the following drawbacks: low detection efficiency, high error rate, limited lenses can be inspected per day, time-consuming, highly subjective, and naked eye damage caused by long-term work under high intensity light source \cite{ding2020automatic} \cite{yang2023deep}. Thus, the replacement of traditional manual detection for machine vision is the inevitable result of industrial development. (Yang et al., 2023).
To address the limitations of the above solutions, deep learning can be utilized in particular computer vision for defect detection. The development of optical lens defect detection system, coupled with deployment of the model on edge device to detect defects in real time can lead to an optimized solution in terms of cost, coverage and real time detection.
This study presents a deep learning-based defect detection system utilizing YOLOv8, trained on a custom dataset of optical lens images. The proposed system aims to provide high-speed, high-accuracy defect detection suitable for real-time industrial applications. The research explores the effectiveness of YOLOv8 in identifying defects and compares its performance to existing techniques. The findings contribute to the advancement of automated optical inspection, offering a scalable and reliable solution for quality control in lens manufacturing.

\section{Literature Review} 
There has been limited research on automated defect detection in optical lenses. Existing studies predominantly focus on two key approaches: traditional image processing techniques and deep learning-based object detection models. Image processing techniques often leverage specialized lighting setups, such as fringe deflectometry and dark-field illumination, to enhance defect visibility. Deep learning approaches, on the other hand, aim to automate feature extraction and improve detection accuracy through advanced neural networks. While both methods have demonstrate effectiveness, they have various trade-offs in terms of accuracy, computational complexity, and real time application.

Starting with image processing techniques, Pan et al. \cite{pan2019comprehensive} proposed a comprehensive, automated defect-detection system for small-sized curved optical lenses using fringe deflectometry, dark-field illumination, and light transmission. Their system achieved an impressive detection accuracy of up to 6 µm for three major defect types: surface-profile defects, internal impurities, and haze defects. However, while their system provided high precision, its reliance on specialized lighting conditions and controlled environments raises concerns about scalability in real-world factory settings.

Similarly, Ding et al. \cite{ding2020automatic} developed a machine vision-based system for detecting dispersed defects in resin eyeglasses. Their approach combined transmission and reflection imaging to enhance gray-scale gradient contrasts in defect regions. A specialized image processing algorithm was then applied for segmentation and enhancement. The system achieved a detection accuracy of 97.5\% with an average processing time of 0.636 seconds per lens, making it well-suited for online inspection systems. However, this method's dependence on predefined lighting conditions and handcrafted features limits its adaptability to lenses of varying shapes, materials, and defect types.

Xian Tao et al. [7] \cite{tao2015novel} extended the use of bright-field and dark-field illumination techniques to inspect large-aperture optical elements. Their system could scan 810 mm × 460 mm lenses in under six minutes, effectively detecting scratches, pits, and dust. However, it was incapable of identifying fine surface watermarks, which are common defects in smaller, curved optical lenses. Additionally, the mechanical setup and operational constraints made the method impractical for high-speed production lines.

Focusing on limited literature specific to machine learning techniques for defect detection in optical lenses, Yang et al. \cite{yang2023deep} introduced MVIS (Micro Vision Inspection System) combined with an enhanced deep-learning model (ISE-YOLO) to detect weak micro-defects on optical lenses. Their model incorporated an Improved Squeeze-and-Excitation (ISE) module and a PolyLoss function to correct class imbalance and enhance feature extraction. Experimental results showed that ISE-YOLO outperformed Faster R-CNN, YOLOv3, YOLOv5, YOLOv6, and YOLOv7, achieving an mAP of 94.23\% and an F1 score of 90.60\%. However, despite its strong performance, ISE-YOLO’s computational cost was not analyzed, raising questions about its feasibility for real-time defect detection on edge devices.

Tang et al. \cite{tang2023improved} proposed STMask R-CNN, a deep-learning approach designed to detect low-contrast, varying-size, and overlapping defects in optical lenses. By incorporating a Swin Transformer, the model improved feature representation and detection accuracy. Using a dataset of 3,800 images with five different defect types, STMask R-CNN outperformed SSD, Faster R-CNN, RetinaNet, and YOLOv5, achieving 98.2\% precision, 97.7\% recall, and an mAP@0.5 of 98.1\%. However, Mask R-CNN-based architectures are computationally expensive, making them less practical for real-time deployment in industrial environments.

Lin et al. \cite{lin2024optical} introduced WGSO-YOLO, an improved optical lens defect detection model that builds upon the YOLO framework. Their approach incorporates three key enhancements: GSConv (a lightweight convolution module) to reduce computational complexity, SOCA (Second-Order Channel Attention) to enhance feature extraction, and WIoU (Wise-IoU loss function) to improve localization accuracy and handle low-quality samples more effectively. The model was evaluated on a custom optical lens defect dataset containing 1,059 images and achieved a mean average precision (mAP@0.5) of 0.927 with a processing speed of 96 FPS. The performance on unseen defect types and real-world variations (e.g., lighting changes, motion blur) remains unclear.

While the image processing techniques reviewed provide high detection accuracy, they are heavily dependent on controlled environments and specific lighting conditions, limiting their adaptability in dynamic manufacturing settings. These methods also struggle with generalizing to different lens types and defect variations.On the other hand, deep learning-based models have shown promising improvements in detection accuracy and robustness. Most studies focus primarily on accuracy metrics (mAP, precision, recall) without considering computational feasibility for real-time industrial deployment.

Considering these limitations, this study adopts YOLOv8 for optical lens defect detection, as it offers a balanced trade-off between accuracy, speed, and computational efficiency. Compared to previous YOLO versions, YOLOv8 features an improved backbone, advanced anchor-free detection, and better feature aggregation, making it well-suited for real-time industrial deployment. Furthermore, YOLOv8’s lighter architecture allows it to be deployed on edge computing devices, enabling real-time defect detection in factory environments.

\section{Methodology}

\subsection{Data Acquisition}
To the best of our understanding, there is no open-source optical lens dataset consisting of representative samples that can be used for developing automated defect detection system for optical lenses. Hence, a dataset was created to address this primary requirement. The first step in constructing the dataset involved capturing images of optical lenses. Each lens was photographed using a high-resolution digital camera under controlled lighting conditions to minimize glare and reflections. This setup ensured that the images were uniform in quality and that variations in environmental factors did not compromise the defect detection process. Once captured, the images were examined for clarity, and any blurry or low-resolution samples were discarded to maintain a high standard of input data.
The proposed architecture would be deployed onto an edge device, the camera connected to edge device would be strategically placed in such a manner that it scan lenses moving on a conveyor belt. Hence, to model this scenario, the images were captured from the top.

\begin{figure}[h]
    \centering
    \includegraphics[width=0.7\textwidth]{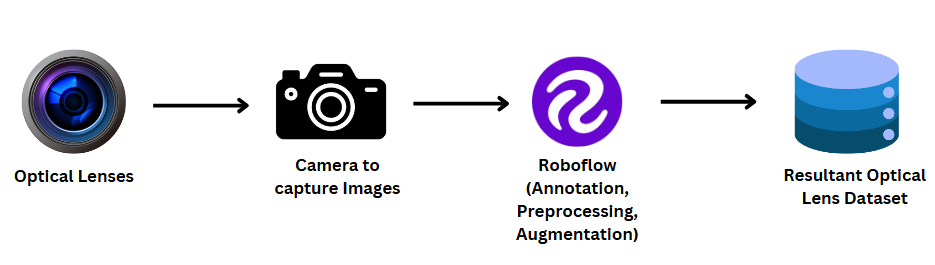} 
    \caption{Dataset Creation} 
    \label{fig:sample_figure} 
\end{figure}

\subsection{Data Annotation}
After collecting the images, the next step involved the annotating of the data. The images were annotated via Roboflow. Annotation primarily involved drawing bounding boxes around lenses and other regions of interest that exhibited potential defects, such as scratches, chips, or accumulated debris, and labeling them accordingly. This step required careful attention to detail, as the accuracy of annotations directly influences model performance. Annotations were reviewed multiple times to ensure consistency and correct labeling, ultimately producing a high-quality annotated dataset.

\subsection{Data Preprocessing}
To enhance dataset variability and robustness, several preprocessing techniques were applied. First, images were auto-oriented to correct any misalignment, followed by resizing to a consistent resolution to ensure uniformity across different models. Since the dataset contained images with varying resolutions, black padding was used to preserve the original aspect ratio while standardizing the resolution. This resizing step was essential to improve model efficiency and reduce training time.

\subsection{Data Augmentation}
The next phase within the data transformation involved the scaling of the dataset through representative augmentations in order to provide enough training data for the selected architecture to generalize on.
To enhance the robustness and generalization of the defect detection model, we applied a set of data augmentation techniques tailored to the real-world operational constraints of an optical lens inspection system. Since the camera is fixed above the conveyor belt, augmentations were designed to simulate lighting variations, lens movement inconsistencies, occlusions, and imaging artifacts. These augmentations improve model performance by reducing overfitting and enabling detection under different real-world conditions.
The lenses would move on a conveyor belt and hence could be in any orientation and would not be in the same position on the conveyor belt every time they pass through the camera. To compensate for this variance and enable the model to accurately inference regardless of the orientation rotation, flip and shift augmentations were applied. The rotation of an image with respect to an angle (nnnnnnnnnnnnnnnnnnnnnnnnnn) would be reached via the matrix:
The shift transformation was introduced in the form of (Sx,Sy), which represents the shift in the x and y directions.
As post deployment, the detection will be performed on lenses moving along the conveyor belt, the images captured by camera could be blurry depending on the speed of conveyor belt. To cater for this type of variance, gaussian blur is utilized. The pixel-wise blurring of certain images was aimed at introducing a representative variance; i.e., due to varying lighting conditions and the hardware specifications of camera device, the model was provided with a richer training dataset to assist with generalization.
Enter a figure here after applying respective augmentation.

\subsection{Architecture Selection}
Once the dataset transformation was complete, the next step was the selection of the architecture that would be used for training in the dataset. There are many object detection architectures available for defect detection, however the majority of these require dedicated GPU’s and are usually hosted on the cloud due to the high computational demand. This is due to the deployment of a “two-stage” detection methodology, whereby, the architecture first generates regional proposals and then object classification is carried out on each proposal. Although this does provide a higher degree of accuracy, the increased computational load makes it undeployable at the edge. The purpose of this research was to train a Lightweight architecture that is able to carry out inference within an enclosed environment (raspberry pi) with a high degree of accuracy and in real-time. Hence, we decided to select a “one-shot” detector, carrying out both classification and bounding-box regression in a single stage.

\subsection{Architecture of YOLOv8}
YOLOv8n consists of 225 layers in its full configuration, this includes all convolutional layers, specialized modules, and other operations that collectively form the detection pipeline. However, the fused version of the model has 168 layers, as certain operations (such as batch normalization) are fused to streamline inference and improve efficiency.
\begin{table}[h]
    \centering
    \caption{YOLOv8 Architecture} 
    \label{tab:Y8arch} 
    \renewcommand{\arraystretch}{1.2}
    \begin{tabular}{l c c c c c}
        \toprule
        \textbf{Layer} & \textbf{Activation} & \textbf{Filters} & \textbf{Size} & \textbf{Repeat} & \textbf{Output Size} \\
        \midrule
        Image   & -      & -    & -        & -  & 640 × 640 \\
        Conv0   & SiLU   & 16   & 3 × 3 / 2 & 1  & 320 × 320 \\
        Conv1   & SiLU   & 32   & 3 × 3 / 2 & 1  & 160 × 160 \\
        C2f0    & SiLU   & 32   & 1 × 1     & 1  & 160 × 160 \\
        Conv2   & SiLU   & 64   & 3 × 3 / 2 & 1  & 80 × 80 \\
        C2f1    & SiLU   & 64   & 1 × 1     & 2  & 80 × 80 \\
        Conv3   & SiLU   & 128  & 3 × 3 / 2 & 1  & 40 × 40 \\
        C2f2    & SiLU   & 128  & 1 × 1     & 2  & 40 × 40 \\
        Conv4   & SiLU   & 256  & 3 × 3 / 2 & 1  & 20 × 20 \\
        C2f3    & SiLU   & 256  & 1 × 1     & 1  & 20 × 20 \\
        SPPF    & SiLU   & 256  & -         & 1  & 20 × 20 \\
        Upsample& -      & -    & -         & 1  & 40 × 40 \\
        Concat  & -      & -    & -         & 1  & 40 × 40 \\
        C2f4    & SiLU   & 128  & 1 × 1     & 1  & 40 × 40 \\
        Upsample& -      & -    & -         & 1  & 80 × 80 \\
        Concat  & -      & -    & -         & 1  & 80 × 80 \\
        C2f5    & SiLU   & 64   & 1 × 1     & 1  & 80 × 80 \\
        Conv5   & SiLU   & 64   & 3 × 3 / 2 & 1  & 40 × 40 \\
        Concat  & -      & -    & -         & 1  & 40 × 40 \\
        C2f6    & SiLU   & 128  & 1 × 1     & 1  & 40 × 40 \\
        Conv6   & SiLU   & 128  & 3 × 3 / 2 & 1  & 20 × 20 \\
        Concat  & -      & -    & -         & 1  & 20 × 20 \\
        C2f7    & SiLU   & 256  & 1 × 1     & 1  & 20 × 20 \\
        Detect  & SiLU   & Varies & -       & -  & Varies \\
        \bottomrule
    \end{tabular}
\end{table}

\subsubsection{Backbone: Feature Extraction}
The backbone is an evolved version of CSPDarknet53, incorporating structural modifications for improved efficiency. It retains CSP-style feature reuse but integrates additional optimizations, including the C2f (Cross-Stage Partial Fusion) module, making it lighter and faster than previous CSPDarknet implementations.
The YOLOv8 model begins with a series of convolutional layers that progressively downsample the input image while increasing the number of filters to extract meaningful features. The initial Conv0 layer applies 16 filters, followed by Conv1 with 32 filters, both utilizing SiLU activation to enhance non-linearity. The first C2f (Cross-Stage Partial Fusion) block with 32 filters introduces a lightweight residual structure, improving gradient flow and feature reuse. Another convolutional layer (Conv2) with 64 filters further extracts features, maintaining a balance between spatial resolution and semantic richness.

As the model progresses, additional C2f blocks refine feature hierarchies. A C2f block with 64 filters enhances representation before Conv3 increases the channels to 128, followed by another C2f block at the same level. Conv4 increases the filter count to 256, allowing deeper feature extraction, followed by another C2f block to reinforce learning. This structured backbone ensures efficient feature capturing across multiple layers, making it effective for detecting small and large defects in optical lenses.
\subsubsection{Neck: Feature Aggregation}
To aggregate multi-scale information, the model incorporates the Spatial Pyramid Pooling Fast (SPPF) module. SPPF enhances receptive fields by pooling features at different kernel sizes, improving object localization and detection accuracy. The upsampling layers then increase the spatial resolution of feature maps, which are concatenated with earlier feature representations. This fusion process allows finer details from shallow layers to complement the deeper semantic features.

Additional convolutional layers such as Conv5 and Conv6 refine these fused features, ensuring an optimal balance between high-resolution spatial details and deeper abstract representations. The model then undergoes another round of upsampling, concatenation, and C2f blocks, ensuring that features at different scales are properly integrated, enabling better detection of defects at varying sizes.
\subsubsection{Head: Detection and Prediction}
The detection head is responsible for generating predictions based on the aggregated feature maps. The Detect layer processes multi-scale features to predict bounding boxes and class probabilities for two object categories: "Lens" and "Defect." This multi-scale approach ensures that both small and large defects are detected effectively. The model employs Distribution Focal Loss (DFL) to refine localization accuracy, ensuring precise bounding box regression.
Moreover, the model’s feature pyramid structure enables it to detect defects at different spatial scales by leveraging feature maps from different layers. This combination of a strong backbone, feature aggregation mechanisms, and an efficient detection head allows YOLOv8 to achieve high accuracy while maintaining real-time inference speed, making it suitable for edge computing environments such as the Jetson Nano.

The proposed defect detection method is based on YOLOv8 model to perform inference in real time. Rather than training a model from scratch transfer learning is utilized to train the pre-trained YOLOv8n model to detect defects in optical lenses. The reason behind using this variant of YOLOv8 is domain requirement for a lightweight architecture that could be deployed onto constrained edge devices, demanding less power consumption.

\section{Experimental Results}
\subsection{Evaluation Metrics}
To thoroughly assess the performance of the model, a range of evaluation metrics were utilized, each offering a distinct perspective on its effectiveness. Alongside fundamental performance indicators like precision, recall, and F1-score, the Intersection over Union (IoU) was employed, as it is a crucial metric in object detection tasks. IoU, often referred to as the Jaccard Index, quantifies the extent of agreement between the predicted bounding boxes \( M_p \) and the actual ground truth boxes \( M_g \). This is mathematically expressed as:
\[
IoU = \frac{area(M_p \cap M_g)}{area(M_p \cup M_g)}
\]

For this research, the IoU threshold was set to 0.5, meaning that a predicted bounding box had to overlap by at least 50\% with the corresponding ground truth box to be considered a correct detection. This threshold ensured that the model accurately located defects within the optical lenses.Furthermore, mean average precision (mAP) was selected as a comprehensive metric to evaluate the model's precision across different confidence levels. Precision, recall, and F1-score were initially computed at a 50\% confidence threshold, while mAP was determined as follows:

\[
mAP = \frac{1}{C} \sum_{i=1}^{C} AP_i
\]

Here, \( AP_i \) denotes the average precision for the \(ith\) object class, while \(C\) represents the total number of object classes in the dataset. The incorporation of \(mAP\) allowed for a more nuanced assessment of the model’s ability to correctly identify and localize defects across various difficulty levels.
Additionally, the real-time applicability of the model was measured in terms of inference speed, expressed in frames per second (FPS). Given that the model was designed for deployment on a Jetson Nano within an industrial setting, it was imperative that it processed images swiftly enough to keep up with the conveyor belt movement. Moreover, false positive and false negative rates were examined to determine the model’s reliability in distinguishing between defective and non-defective lenses, ensuring minimal errors in a real-world application.

\subsection{Model Evaluation}
The evaluation metrics reveal that the model detects lenses with high accuracy but struggles with defect identification. Precision, which measures the proportion of correctly identified bounding boxes among all detected objects, is 0.688 overall, with 0.944 for lenses and 0.431 for defects. The lower precision for defects suggests a higher number of false positives, where non-defective areas are incorrectly classified as defects. Recall, which indicates the proportion of actual objects correctly detected, is 0.624 overall, with 1.000 for lenses and only 0.249 for defects. The perfect recall score for lenses suggests that all lenses are being detected, whereas the low recall for defects indicates that many defects are being missed.

\begin{table}[h]
    \centering
    \caption{Evaluation Metrics for Object Detection} 
    \label{tab:evaluation_metrics} 
    \renewcommand{\arraystretch}{1.2}
    \begin{tabular}{l c c c c c c}
        \toprule
        \textbf{Class} & \textbf{Images} & \textbf{Instances} & \textbf{Box (P)} & \textbf{R} & \textbf{mAP50} & \textbf{mAP50-95} \\
        \midrule
        all    & 50  & 172 & 0.688 & 0.624 & 0.617 & 0.545 \\
        defect & 19  & 122 & 0.431 & 0.249 & 0.239 & 0.0957 \\
        lens   & 50  & 50  & 0.944 & 1.000 & 0.995 & 0.995 \\
        \bottomrule
    \end{tabular}
\end{table}

The mean Average Precision (mAP) values further highlight the discrepancy in detection performance between the two classes. The overall mAP@50 (IoU=0.5) is 0.617, and the stricter mAP@50-95 is 0.545. When analyzing individual classes, the model achieves a near-perfect mAP@50-95 of 0.995 for lenses, while the same metric for defects is significantly lower at 0.0957, suggesting that the model struggles to localize and classify defects accurately. This poor performance in defect detection could be attributed to factors such as dataset imbalance, where the number of defect instances is significantly lower than the number of lenses, leading to biased learning. The variability in defect appearance may also play a role, as defects can have diverse shapes, textures, and lighting effects, making them harder to generalize.
Despite the limitations in defect detection, the model maintains an efficient inference speed, processing each image in 0.6ms for preprocessing, 5.7ms for inference, and 1.8ms for postprocessing, making it well-suited for real-time applications.

\begin{figure}[h]
    \centering
    \includegraphics[width=0.7\textwidth]{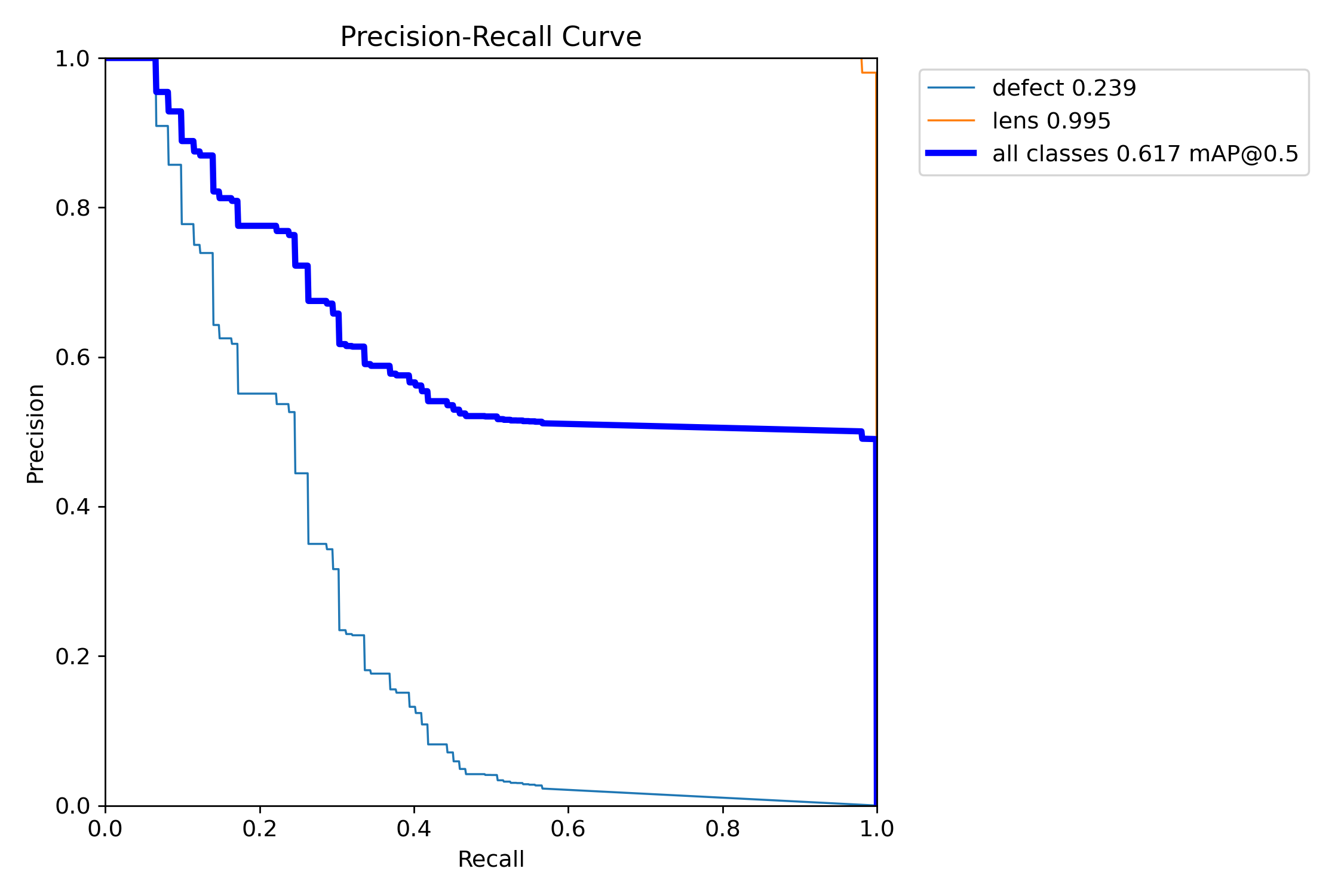} 
    \caption{PR curve} 
    \label{fig:PR curve} 
\end{figure}

Figure \ref{fig:PR curve} illustrates the Precision-Recall (PR) curves for the detection of both lens defects and the lens itself. The model demonstrates strong performance in identifying the lens (mAP@0.5 = 0.995), achieving near-perfect precision and recall. The decline in PR curve for defect class indicates that as the model’s recall increases, its precision decreases, this can be due to the fact that some lens contained hundreds of small sized defects which couldn’t be labeled properly. The model while inference picks on these defects due to which a steep decline in precision can be seen.

\begin{figure}[h]
    \centering
    \includegraphics[width=0.7\textwidth]{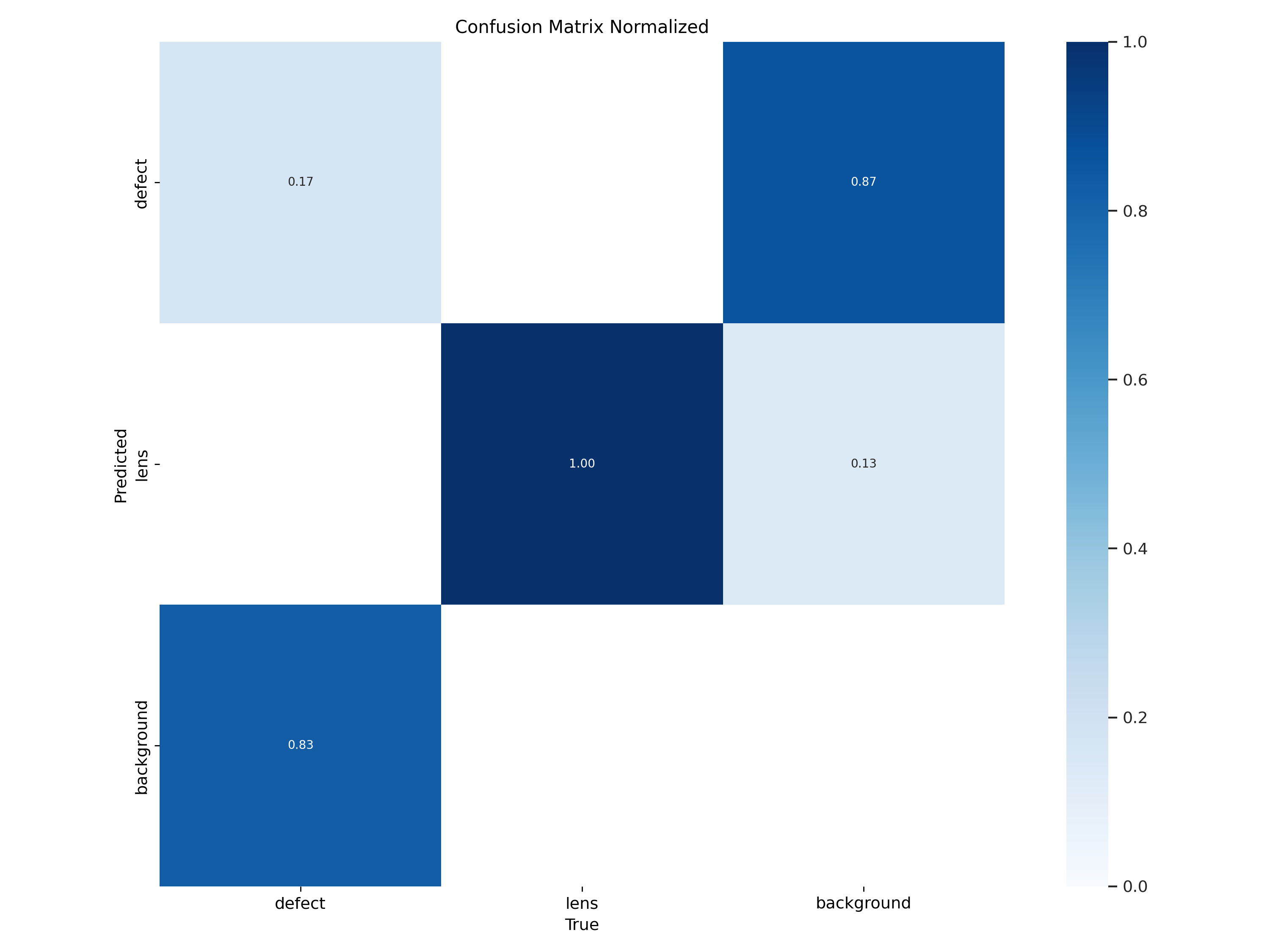} 
    \caption{Confusion Matrix} 
    \label{fig:CFMatrix} 
\end{figure}

This normalized confusion matrix (Figure \ref{fig:CFMatrix}) shows how well the model classifies defects, lenses, and background. The ideal values should be high along the diagonal, meaning correct classifications. The model performs well in detecting lenses, achieving nearly perfect classification with a value of 1.00. However, the high number of false positives and false negatives in defect detection can be attributed to the labeling process. Some lenses contained numerous small defects which could not be labeled. Due to this when the model identifies unlabeled defects, they are represented as false positive and negative in the confusion matrix. This does not necessarily indicate that the model is poor at detecting defects. Addressing this issue may involve refining the annotation strategy to ensure all relevant defects are labeled, thereby improving the model's ability to distinguish between defects, lenses, and background.

\section{Discussion}
The model demonstrates strong performance in detecting lenses, achieving near-perfect classification with an accuracy of 1.00. However, its performance in detecting defects is notably lower, as reflected in the evaluation metrics. The primary reason for this discrepancy lies in the dataset annotation process. Some lenses contained numerous small defects that were not labeled due to annotation limitations. Consequently, when the model detects these unlabeled defects, they are mistakenly represented as false positives or false negatives, which negatively impacts precision and recall values.
Despite the lower defect detection metrics, this does not necessarily indicate that the model is ineffective in identifying defects. Instead, it highlights the challenge of accurately labeling small defects and the impact of annotation quality on model evaluation. The confusion matrix further confirms this, as many detected defects are misclassified due to the absence of ground truth labels. Addressing this issue could involve refining the annotation process by incorporating finer-grained labeling or utilizing automated labeling techniques to reduce inconsistencies.

\section{Conclusion}
This study developed a YOLOv8-based deep learning system for automated optical lens defect detection, addressing the limitations of manual inspection. The system efficiently detects defects making it suitable for real-time industrial use. The model achieved 10 FPS while inference in real time on jetson nano. By leveraging YOLO, it provides a scalable and efficient solution for detecting various defects in optical lenses, improving consistency and reducing the reliance on human inspectors. This approach offers significant potential for enhancing quality control in optical lens manufacturing. Future work could focus on optimizing the model further and expanding its defect detection capabilities, in addition to extending our methodological approach to other domains where computer vision is being applied such as healthcare ~\cite{hussain2023child} and emotion detection ~\cite{aydin2023domain}.

\begin{adjustwidth}{-\extralength}{0cm}

\bibliographystyle{unsrt}  
\bibliography{ref}  

\end{adjustwidth}
\end{document}